\documentclass[11pt,a4paper]{article}
\usepackage[hyperref]{naaclhlt2018}
\usepackage{times}
\usepackage{latexsym}
\usepackage{booktabs}

\usepackage{amsfonts,amsmath,amsthm}

\newcommand{\Beta}{B}

\usepackage{url}

\usepackage{soul,color}
\usepackage{todonotes} %add [disable] to hide todonotes

\aclfinalcopy % Uncomment this line for the final submission
\title{On Evaluating the Generalization of LSTM Models in Formal Languages}

\author{Mirac Suzgun \hfill Yonatan Belinkov \hfill Stuart M. Shieber \\\\ 
  John A. Paulson School of Engineering and Applied Sciences \\ 
  Harvard University  \\
  Cambridge, MA 02138, USA \\
  {\tt \{msuzgun@college,belinkov@seas,shieber@seas\}.harvard.edu} }

\date{}

\begin{document}
\maketitle

\begin{abstract}
Recurrent Neural Networks (RNNs) are theoretically Turing-complete and established themselves as a dominant model for language processing. Yet, there still remains an uncertainty regarding their language learning capabilities. In this paper, we empirically evaluate the inductive learning capabilities of Long Short-Term Memory networks, a popular extension of simple RNNs, to learn simple formal languages, in particular $a^n b^n$, $a^n b^n c^n$, and $a^n b^n c^n d^n$. We investigate the influence of various aspects of learning, such as training data regimes and model capacity, on the generalization to unobserved samples. We find striking differences in model performances under different training settings and highlight the need for careful analysis and assessment when making claims about the learning capabilities of neural network models.\footnote{Our code is available at \url{https://github.com/suzgunmirac/lstm-eval}.}  
\end{abstract}

\section{Introduction}
Recurrent Neural Networks (RNNs) are powerful machine learning models that can capture and exploit sequential data. They have become standard in important natural language processing tasks such as machine translation \citep{sutskever2014sequence,bahdanau2014neural} and speech recognition \citep{DBLP:conf/interspeech/SakSB14}. Despite the ubiquity of various RNN architectures in natural language processing, there still lies an unanswered fundamental question: What classes of languages can, empirically or theoretically, be learned by neural networks? This question has drawn much attention in the study of formal languages, with previous results on both the theoretical \citep{siegelmann1992computational, siegelmann1995computation} and empirical capabilities of RNNs, showing that different RNN architectures can learn certain regular \citep{giles1992learning, casey1996dynamics}, context-free \citep{elman1991distributed,das1992learning}, and context-sensitive languages \citep{gers2001lstm}. 

In a common experimental setup for investigating whether a neural network can learn a formal language, one formulates a supervised learning problem %platform
where the network is presented one character at a time and predicts the next possible character(s). The performance of the network can then be evaluated based on its ability to recognize sequences shown in the training set and -- more importantly -- to generalize to unseen sequences. There are, however, various methods of evaluation in a language learning task. In order to define the \textit{generalization} of a network, one may consider the length of the shortest sequence in a language whose output was incorrectly produced by the network, or the size of the largest accepted test set, or the accuracy on a fixed test set \citep{Rodriguez99arecurrent,boden2000context, gers2001lstm,rodriguez2001simple}. These formulations follow narrow and bounded evaluation schemes though: They often define a length threshold in the test set and report the performance of the model on this fixed set.

We acknowledge three unsettling issues with these formulations. First, the sequences in the training set are usually assumed to be uniformly or geometrically distributed, with little regard to the nature and complexity of the language. This assumption may undermine any conclusions drawn from empirical investigations, especially given that natural language is not uniformly distributed, an aspect that is known to affect learning in modern RNN architectures \citep{liu-levy-schwartz-tan-smith:2018:RepL4NLP}. Second, in a test set where the sequences are enumerated by their lengths, if a network makes an error on a sequence of, say, length $7$, but correctly recognizes longer sequences of length up to $1000$, would we consider the model's generalization as good or bad? In a setting where we monitor only the shortest sequence that was incorrectly predicted by the network, this scheme clearly misses the potential success of the model after witnessing a failure, thereby misportraying the capabilities of the network. Third, the test sets are often bounded in these formulations, making it challenging to compare and contrast the performance of models if they attain full accuracy on their fixed test sets. 

In the present work, we address these limitations by providing a more nuanced evaluation of the learning capabilities of RNNs. In particular, we investigate the effects of three different aspects of a network's generalization: data  distribution, length-window, and network capacity. We define an informative protocol for assessing the performance of RNNs: Instead of training a single network until it has learned its training set and then evaluating it on its test set, as \citeauthor{gers2001lstm} do in their study, we monitor and test the network's performance at each epoch during the entire course of training. This approach allows us to study the stability of the solutions reached by the network. Furthermore, we do not restrict ourselves to a test set of sequences of fixed lengths during testing. Rather, we exhaustively enumerate all the sequences in a language by their lengths and then go through the sequences in the test set one by one until our network errs $k$ times, thereby providing a more fine-grained evaluation criterion of its generalization capabilities. 

Our experimental evaluation is focused on the Long Short-Term Memory (LSTM) network \citep{hochreiter1997long}, a particularly popular RNN variant. We consider three formal languages, namely $a^n b^n$, $a^n b^n c^n$, and $a^n b^n c^n d^n$, and investigate how LSTM networks learn these languages under different training regimes. Our investigation leads to the following insights: (1) The data distribution has a significant effect on generalization capability, with discrete uniform and U-shaped distributions often leading to the best generalization amongst all the four distributions in consideration. (2) Widening the training length-window, naturally, enables LSTM models to generalize better to longer sequences, and interestingly, the networks seem to learn to generalize to shorter sequences when trained on long sequences. (3) Higher model capacity -- having more hidden units -- leads to better stability, but not necessarily better generalization levels. In other words, over-parameterized models are more stable than models with theoretically sufficient but far fewer parameters. We explain this phenomenon by conjecturing that a collaborative counting mechanism arises in over-parameterized networks.

\section{Related Work}

% \begin{table*}[t]
% \centering 
% \begin{tabular}{c | c | c} 
% \toprule
% \textbf{Formal Language} & \textbf{Automaton} & \textbf{Examples} \\  
% \midrule
% Regular & Finite State Automaton & $a^5 b^7, a^* b^*, a^* bc d^*$\\
% \midrule
% Context-Free & Pushdown Automaton & $a^n b^n$, $w\# w^R$, $a^n b^m c^n d^m$ \\
% \midrule
% Context-Sensitive & Linearly Bounded Automaton & $a^n b^n c^n$, $a^{2^n}$, $a^p$ (where $p$ is prime) \\
% \midrule
% Recursively Enumerable & Turing Machine  & Entscheidungsproblem, Mortality problem\\
% \bottomrule 
% \end{tabular}
% \caption{The correspondence between formal languages and abstract automata  \citep{jager2012formal}}
% \label{table:chomsky} 
% \end{table*}

It has been shown that RNNs with a finite number of states can process regular languages by acting like a finite-state automaton using different units in their hidden layers \citep{giles1992learning, casey1996dynamics}. RNNs, however, are not limited to recognizing only regular languages. \citet{siegelmann1992computational} and \citet{siegelmann1995computation} showed that first-order RNNs (with rational state weights and infinite numeric precision) can simulate a pushdown automaton with two-stacks, thereby demonstrating that RNNs are Turing-complete. In theory, RNNs with infinite numeric precision are capable of expressing recursively enumerable languages. Yet, in practice, modern machine architectures do not contain computational structures that support infinite numeric precision. Thus, the computational power of RNNs with finite precision may not necessarily be the same as that of RNNs with infinite precision.

\citet{elman1991distributed} investigated the learning capabilities of simple RNNs to process and formalize a context-free grammar containing hierarchical (recursively embedded) dependencies: He observed that distinct parts of the networks were able to  learn some complex representations to encode certain grammatical structures and dependencies of the context-free grammar. Later, \citet{das1992learning} introduced an RNN with an external stack memory to learn simple context-free languages, such as $a^n b^m$,  $a^nb^ncb^ma^m$, and $a^{n+m} b^n c^m$. Similar studies \citep{kwasny1995tail, wiles1995learning, steijvers1996recurrent, Rodriguez99arecurrent, boden2000context} have explored the existence of stable counting mechanisms in simple RNNs, which would enable them to learn various context-free and context-sensitive languages, but none of the RNN architectures proposed in the early days were able to generalize the training set to longer (or more complex) test samples with substantially high accuracy.

\begin{table*}[t]
\centering
\begin{tabular}{l | cccc | cccccc | cccccccc}
\toprule
 \bf Sample & \multicolumn{4}{c|}{$a^2 b^2$} & \multicolumn{6}{c|}{$ a^2 b^2 c^2$} & \multicolumn{8}{c}{$ a^2 b^2 c^2 d^2$} \\ 
 \midrule
 %$n$ & 4 & 3 & 2 \\ 
%\bf  Vocabulary & \multicolumn{4}{c|}{\{a,b\}} & \multicolumn{6}{c|}{\{a,b,c\}} & \multicolumn{8}{c}{\{a,b,c,d\}} \\ 
 \bf  Input &  $a$&$a$&$b$&$b$  & $a$&$a$&$b$&$b$&$c$&$c$ & $ a$&$a$&$b$&$b$&$c$&$c$&$d$&$d $ \\ 
 \bf Output & $a/b$&$a/b$&$b$&$\dashv$ & $a/b$&$a/b$&$b$&$c$&$c$&$\dashv $ & $ a/b$&$a/b$&$b$&$c$&$c$&$d$&$d$&$\dashv $ \\ 
\bottomrule
\end{tabular}
\caption{Example input-output pairs for each language under the sequence prediction formulation.}
\label{tab:examples}
\end{table*}

\citet{gers2001lstm}, on the other hand, proposed a variant of Long Short-Term Memory (LSTM) networks\footnote{For a comprehensive investigation of the LSTM architectures, we invite the reader to refer to the following two papers: \citep{hochreiter1997long, greff2017lstm}.} to learn two context-free languages, $a^n b^n$, $a^n b^m B^m A^n$, and one strictly context-sensitive language, $a^n b^n c^n$. Given only a small fraction of samples in a formal language, with values of $n$ (and $m$) ranging from $1$ to a certain training threshold $N$, they trained an LSTM model until its full convergence on the training set and then tested it on a more generalized set. They showed that their LSTM model outperformed the previous approaches in capturing and generalizing the aforementioned formal languages. By analyzing the cell states and the activations of the gates in their LSTM model, they further demonstrated that the network learns how to count up and down at certain places in the sample sequences to encode information about the underlying structure of each of these formal languages. 

Following this approach, \citet{boden2002learning} and \citet{chalup2003incremental} studied the stability of the LSTM networks in learning context-free and context-sensitive languages and examined the processing mechanism developed by the hidden states during the training phase. They  observed that the weight initialization of the hidden states in the LSTM network had a significant effect on the inductive capabilities of the model and that the solutions were often unstable in the sense that the numbers up to which the LSTM models were able to generalize using the training dataset sporadically oscillated. 

\section{The Sequence Prediction Task}

Following the traditional approach adopted by \citet{elman1991distributed, rodriguez2001simple, gers2001lstm} and many other studies, we train our neural network as follows. At each time step, we present one input character to our model and then ask it to predict the set of next possible characters, based on the current character and the prior hidden states.% weights.
\footnote{Unlike \citet{gers2001lstm}, we do not start each input sequence with a start symbol, since we observed that having a start symbol in the sequence does not affect the learning capabilities of the model. However, we still use a termination symbol $\dashv$ to encode the end of the sequence in our output samples.} Given a vocabulary $\mathcal{V}^{(i)}$ of size $d$, we use a one-hot representation to encode the input values; therefore, all the input vectors are $d$-dimensional binary vectors. The output values are $(d+1)$-dimensional though, since they may further contain the termination symbol $\dashv$, in addition to the symbols in $\mathcal{V}^{(i)}$. The output values are not always one-hot encoded, because there can be multiple possibilities for the next character in the sequence, therefore we instead use a $k$-hot representation to encode the output values. Our objective is to minimize the mean-squared error (MSE) of the sequence predictions. During testing, we use an output threshold criterion of $0.5$ for the sigmoid output layer to indicate which characters were predicted by the model. We then turn this prediction task into a classification task by \textit{accepting} a sample if our model predicts \emph{all} of its output values correctly and \textit{rejecting} it otherwise.\footnote{We note that we only present positive samples from a given language to our model, but this approach is still consistent with Gold's Theorem about the inductive interference of formal languages only from positive samples \citep{angluin1980inductive}, because we give feedback to our model during training whenever it makes an error about its predictions.}

 \subsection{Languages}
We consider the following three formal languages in our predictions tasks: $a^n b^n$, $a^n b^n c^n$, and $a^n b^n c^n d^n$, where $n \geq 1$. Of these three languages, the first one is a context-free language and the last two are strictly context-sensitive languages. Table~\ref{tab:examples} provides example input-output pairs for these languages under the sequence prediction task. In the rest of this section, we formulate the sequence prediction task for each language in more detail. 

% \begin{table*}[t]
% \centering
% \begin{tabular}{c | c | c}
% \toprule
% \bf Language & \bf Input & \bf Output \\ 
% \midrule
% $a^4 b^4$ & $a~a~a~a~b~b~b~b$    & $ a/b~a/b~a/b~a/b~b~b~b~T $   \\ 
% $ a^3 b^2 c^2$ & $ a~a~a~b~b~b~c~c~c $ & $ a/b~a/b~a/b~b~b~c~c~c~T $ \\
% $ a^2 b^2 c^2 d^2 $ & $ a~a~b~b~c~c~d~d $ & $ a/b~a/b~b~c~c~d~d~T $ \\ 
% \bottomrule
% \end{tabular}
% \caption{Example input-output pairs for each language under the sequence prediction formulation.}
% \label{tab:examples}
% \end{table*}

\paragraph{CFL $a^n b^n$:}

The input vocabulary $\mathcal{V}^{(i)}$ for $a^n b^n$ consists of $a$ and $b$. The output vocabulary $\mathcal{V}^{(o)}$ is the union of $\mathcal{V}^{(i)}$ and $\{\dashv\}$. Therefore, the input vectors are $2$-dimensional, and the output vectors are $3$-dimensional. Before the occurrence of the first $b$ in a sequence, the model always predicts $a$ or $b$ (which we notate $a/b$) whenever it sees an $a$. However, after it encounters the first $b$, the rest of the sequence becomes entirely deterministic: Assuming that the model observes $n$ $a$'s in a sequence, it outputs $(n-1)$ $b$'s for the next $(n-1)$ $b$'s and the terminal symbol $\dashv$ for the last $b$ in the sequence. Summarizing, we define the input-target scheme for $a^n b^n$ as follows:
\begin{equation}
    a^n b^n \Rightarrow (a/b)^n b^{n-1}\dashv
\end{equation}

\paragraph{CSL $a^n b^n c^n$:}

The input vocabulary $\mathcal{V}^{(i)}$ for $a^n b^n c^n$ consists of three characters: $a$, $b$, and $c$. The output vocabulary $\mathcal{V}^{(o)}$ is $\mathcal{V}^{(i)} \cup \{\dashv\}$. The input and output vectors are $3$- and $4$-dimensional, respectively. The input-target scheme for $a^n b^n c^n$ is:
\begin{equation}
    a^n b^n c^n\Rightarrow (a/b)^{n}b^{n-1}c^{n}\dashv
\end{equation}

\paragraph{CSL $a^n b^n c^n d^n$:}
The vocabulary $\mathcal{V}^{(i)}$ for the last language $a^n b^n c^n d^n$ consists of $a$, $b$, $c$, and $d$. The input vectors are $4$-dimensional, and the output vectors are $5$-dimensional. As in the case of the previous two languages, a sequence becomes entirely deterministic after the observance of the first $b$, hence the input-target scheme for $a^n b^n c^n d^n$ is:
\begin{equation}
    a^n b^n c^n d^n\Rightarrow (a/b)^n b^{n-1} c^n d^{n}\dashv
\end{equation}

 \subsection{The LSTM Model}
\begin{figure}[t]
    \centering
    \includegraphics[width=0.45\textwidth]{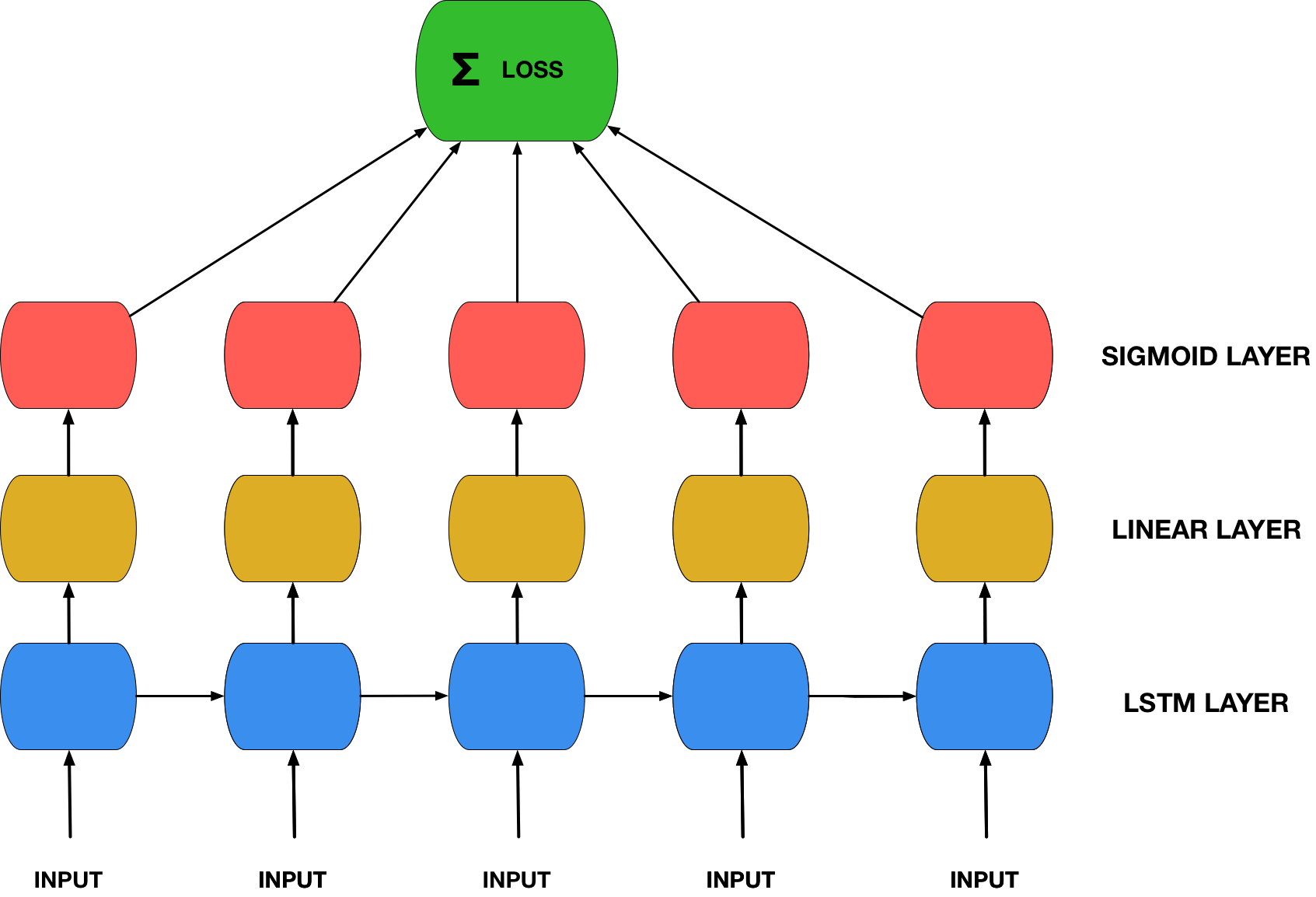}
    \caption{Our LSTM architecture}
    \label{fig:lstm_arch}
\end{figure}

We use a single-layer LSTM model to perform the sequence prediction task, followed by a linear layer   that maps to the output vocabulary size. The linear layer is followed by a sigmoid unit layer. The loss is the sum of the mean squared error between the prediction and the correct output at each character. 
See Figure \ref{fig:lstm_arch} for an illustration.  In our implementation, we used the standard LSTM module in PyTorch \citep{paszke2017automatic} and initialized the initial hidden and cell states, $h_0$ and $c_0$, to zero.

% \begin{align*}
%     i_t &= \sigma(W_{ii} x_t + b_{ii} + W_{hi} h_{(t-1)} + b_{hi}) \\
%     f_t &= \sigma(W_{if} x_t + b_{if} + W_{hf} h_{(t-1)} + b_{hf}) \\
%     g_t &= \tanh(W_{ig} x_t + b_{ig} + W_{hg} h_{(t-1)} + b_{hg}) \\
%     o_t &= \sigma(W_{io} x_t + b_{io} + W_{ho} h_{(t-1)} + b_{ho}) \\
%     c_t &= f_t c_{(t-1)} + i_t g_t \\
%     h_t &= o_t \tanh(c_t)
% \end{align*}

\section{Experimental Setup}

 \subsection{Training and Testing}
Training and testing are done in alternating steps: In each epoch, for training, we first present to an LSTM network $1000$ samples in a given language, which are generated according to a certain discrete probability distribution supported on a closed finite interval.\footnote{The strings are presented to the model in a random order.} We then freeze all the weights in our model, exhaustively enumerate all the sequences in the language by their lengths, and determine the first $k$ shortest sequences whose outputs the model produces inaccurately.\footnote{In all our experiments, we decided to choose $k$ to be $5$.} We remark, for the sake of clarity, that our test design is slightly different from the traditional testing approaches used by \citet{Rodriguez99arecurrent,gers2001lstm,rodriguez2001simple}, since we do not consider the shortest sequence in a language whose output was incorrectly predicted by the model, or the largest accepted test set, or the accuracy of the model on a fixed test set.

\begin{figure*}[t]
\centering
{\includegraphics[width=0.25\textwidth]{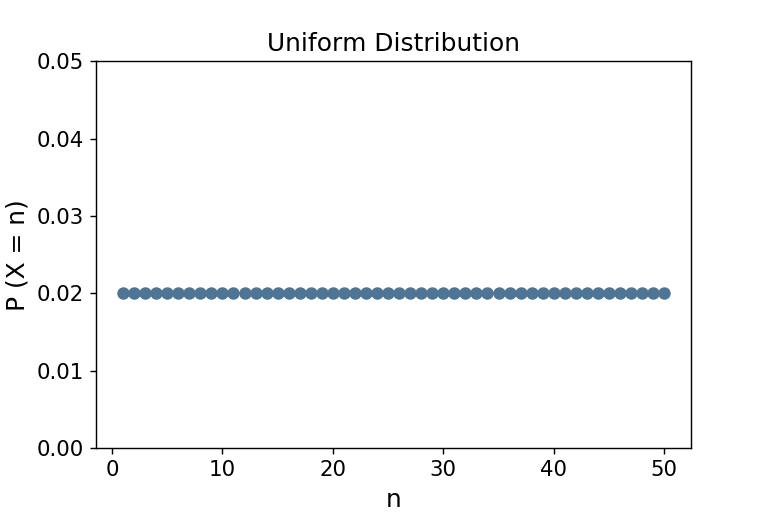}}%
{\includegraphics[width=0.25\textwidth]{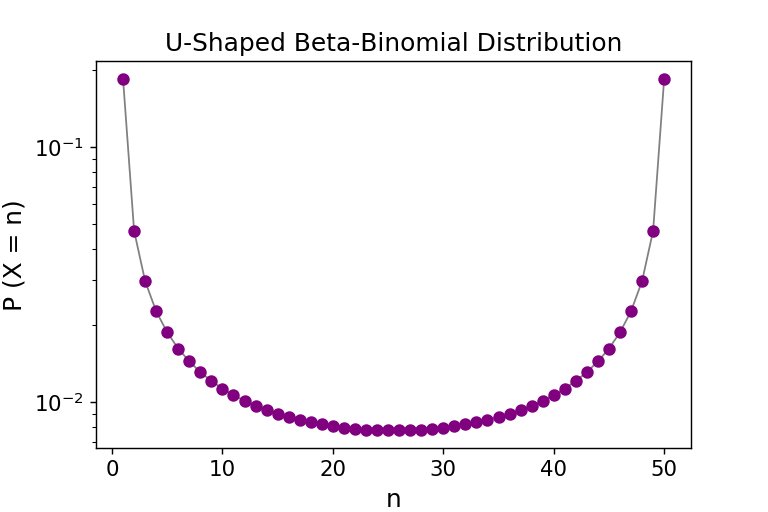}}%
{\includegraphics[width=0.25\textwidth]{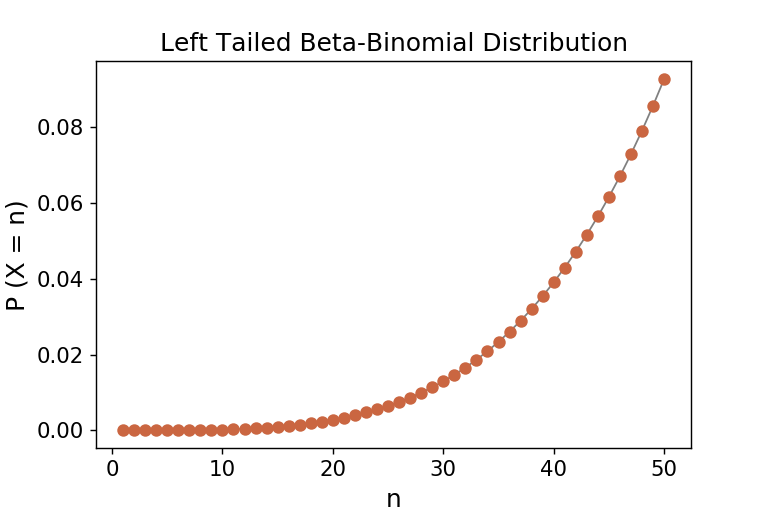}}%
{\includegraphics[width=0.25\textwidth]{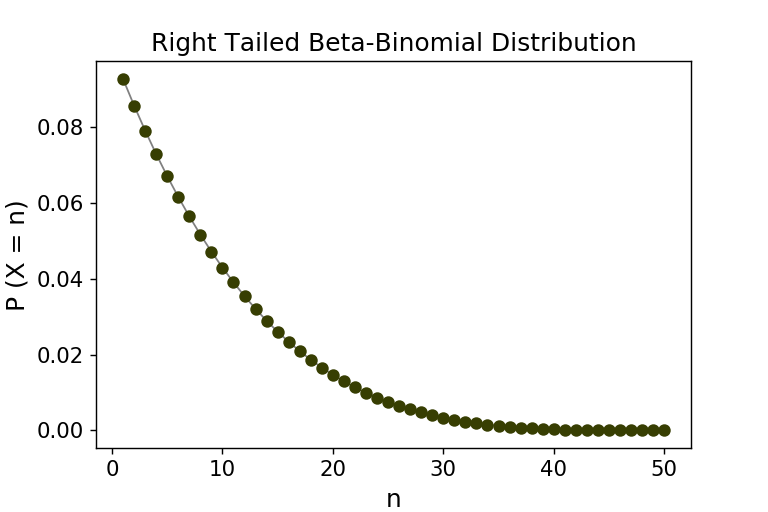}}%
\caption{Distributions from Left to Right: Uniform Distribution with $N=50$, (U-Shaped) Beta-Binomial Distribution with $\alpha = 0.25, \beta = 0.25, N = 49$,  (Right-Tailed) Beta-Binomial Distribution with $\alpha = 1, \beta = 5, N = 49$, and (Left-Tailed) Beta-Binomial Distribution with $\alpha = 5, \beta = 1, N = 49$.}
\label{fig:differentdistributions}
\end{figure*}

Our testing approach, as we will see shortly in the following subsections, gives more information about the inductive capabilities of our LSTM networks than the previous techniques and proves itself to be useful especially in the cases where the distribution of the length of our training dataset is skewed towards one of the boundaries of the distribution's support. For instance, LSTM models sometimes fail to capture some of the short sequences in a language during the testing phase\footnote{This phenomenon is usually observed in distributions where the training set is skewed towards having more long sequences than short sequences.}, but they then predict a large number of long sequences correctly.\footnote{We note that correctly predicting the outputs for the samples $ab$, $abc$, and $abcd$ in the languages $a^n b^n$, $a^n b^n c^n$, and $a^n b^n c^n d^n$, respectively, is a hard task, because the output sequences for these samples are $a\dashv$, $ac\dashv$, and $acd\dashv$, in this given order. While they never contain the symbol $b$ in their outputs, the rest of the sequences in their corresponding languages do contain at least one $b$ in their outputs.} If we were to report only the shortest sequence whose output our model incorrectly predicts, we would then be unable to capture the model's inductive capabilities. Furthermore, we test and report the performance of the model after each full pass of the training set. Finally, in all our investigations, we repeated each experiment ten times. In each trial, we only changed the weights of the hidden states of the model -- all the other parameters were kept the same. 

 \subsection{Length Distributions}

Previous studies have examined various length distribution models to generate appropriate training sets for each formal language: \citet{wiles1995learning, boden2000context, rodriguez2001simple}, for instance, used length distributions that were skewed towards having more short sequences than long sequences given a training length-window, whereas \citet{gers2001lstm} used a uniform distribution scheme to generate their training sets. The latter briefly comment that the distribution of lengths of sequences in the training set does influence the generalization ability and convergence speed\footnote{We define \textit{convergence (learning) speed} as the speed at which a sequence of numbers, the $e_1$ or $e_5$ values in our cases, converge to its stationary value.} of neural networks, and mention that training sets containing abundant numbers of both short and long sequences are learned by networks much more quickly than uniformly distributed regimes. Nevertheless, they do not systematically compare or explicitly report their findings. To study the effect of various length distributions on the learning capability and speed of LSTM models, we experimented with four discrete probability distributions supported on bounded intervals (Figure \ref{fig:differentdistributions}) to sample the lengths of sequences for the languages. % $a^n b^n$, $a^n b^n c^n$, and $a^n b^n c^n d^n$. 
We briefly recall the probability distribution functions for discrete uniform and Beta-Binomial distributions used in our data generation procedure. 

\paragraph{Discrete Uniform Distribution:} Given $N \in \mathbb{N}$, if a random variable $X \sim U (1, N)$, then the probability distribution function of $X$ is given as follows: 
\begin{equation*}
P(x) = 
\begin{cases}
\frac{1}{N} & \text{if } x \in \{1, \ldots, N\} \\ 
0 & \text{otherwise.}
\end{cases}
\end{equation*}
To generate training data with uniformly distributed lengths, we simply draw $n$ from $U (1, N)$ as defined above.  

\paragraph{Beta-Binomial Distribution:} Similarly, given $N \in \mathbb{Z}^{\geq 0}$ and two parameters $\alpha$ and $ \beta \in \mathbb{R}^{>0}$, if a random variable $X \sim \text{BetaBin} (N, \alpha, \beta)$, then the probability distribution function of $X$ is given as follows: 
\begin{equation*}
P(x) = 
\begin{cases}
\binom{N}{x} \frac{\Beta (x+\alpha, N-x+\beta)}{\Beta(\alpha, \beta)} & \text{if } x \in \{0, \ldots, N\} \\ 
0 & \text{otherwise.}
\end{cases}
\end{equation*}
\noindent where $\Beta (\alpha, \beta)$ is the Beta function. We set different values of $\alpha$ and $\beta$ as such in order to generate the following distributions: 

\textbf{U-shaped} ($\alpha = 0.25$, $\beta = 0.25$): The probabilities of having short and long sequences are equally high, but the probability of having an average-length sequence is low.

\textbf{Right-tailed} ($\alpha = 1$, $\beta = 5$): Short sequences are more probable than long sequences. 

\textbf{Left-tailed} ($\alpha = 5$,  $\beta = 1$): Long sequences are more probable than short sequences.

\begin{figure*}[t]
\centering
{\includegraphics[width=0.99\textwidth]{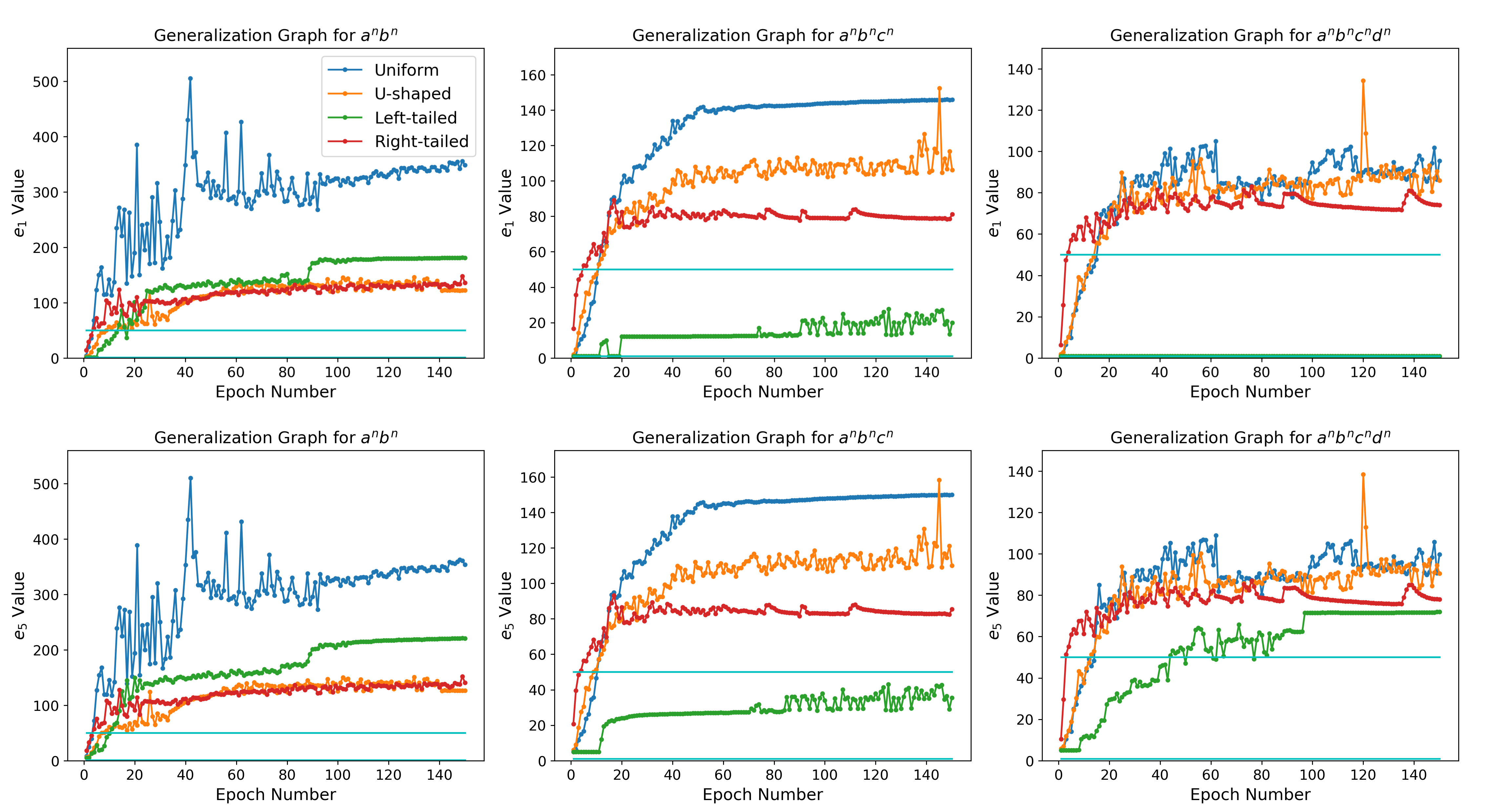}}
%%%%%%%
\caption{Generalization graphs showing the average performance of LSTMs trained under different probability distribution regimes for each language. The top plots show the $e_1$ values, whereas the bottom ones the $e_5$ values. The light blue horizontal lines indicate the training length window $[1, 50]$.}
\label{fig:distributiongraphs}
\end{figure*}

 \subsection{%Sampling from Different 
 Length Windows}
Most of the previous studies trained networks on sequences of lengths $n \in [1, N]$, where typical $N$ values were between 10 and 50 \citep{boden2000context,gers2001lstm}, and more recently 100 \citep{P18-2117}. To determine the impact of the choice of training length-window on the stability and inductive capabilities of the LSTM networks, we experimented with three different length-windows for $n$: $[1, 30]$, $[1, 50]$, and $[50, 100]$. In the third window setting $[50, 100]$, we further wanted to see whether LSTM are capable of generalizing to short sequences that are contained in the window range $[1, 50]$, as well as to sequences that are longer than the sequences seen in the training set.

 \subsection{Model Capacity}
It has been shown by \citet{gers2001lstm} that LSTMs can learn $a^n b^n$ and $a^n b^n c^n$ with $1$ and $2$ hidden units, respectively. Similarly, \citet{holldobler1997designing} demonstrated that a simple RNN architecture containing a single hidden unit with carefully tuned parameters can develop a canonical linear counting mechanism to recognize the simple context-free language $a^n b^n$, for $n \leq 250$. We wanted to explore whether the stability of the networks would improve with an increase in capacity of the LSTM model. We, therefore, varied the number of hidden units in our %single-layer 
LSTM models as follows. We experimented with $1$, $2$, $3$, and $36$ hidden units for $a^n b^n$; $2$, $3$, $4$, and $36$ hidden units for $a^n b^n c^n$; and $3$, $4$, $5$, and $36$ hidden units for $a^n b^n c^n d^n$. The $36$ hidden unit case represents an over-parameterized network with more than enough theoretical capacity to recognize all these languages.  
\section{Results}

\begin{figure*}[h]
\centering
{\includegraphics[width=0.99\textwidth]{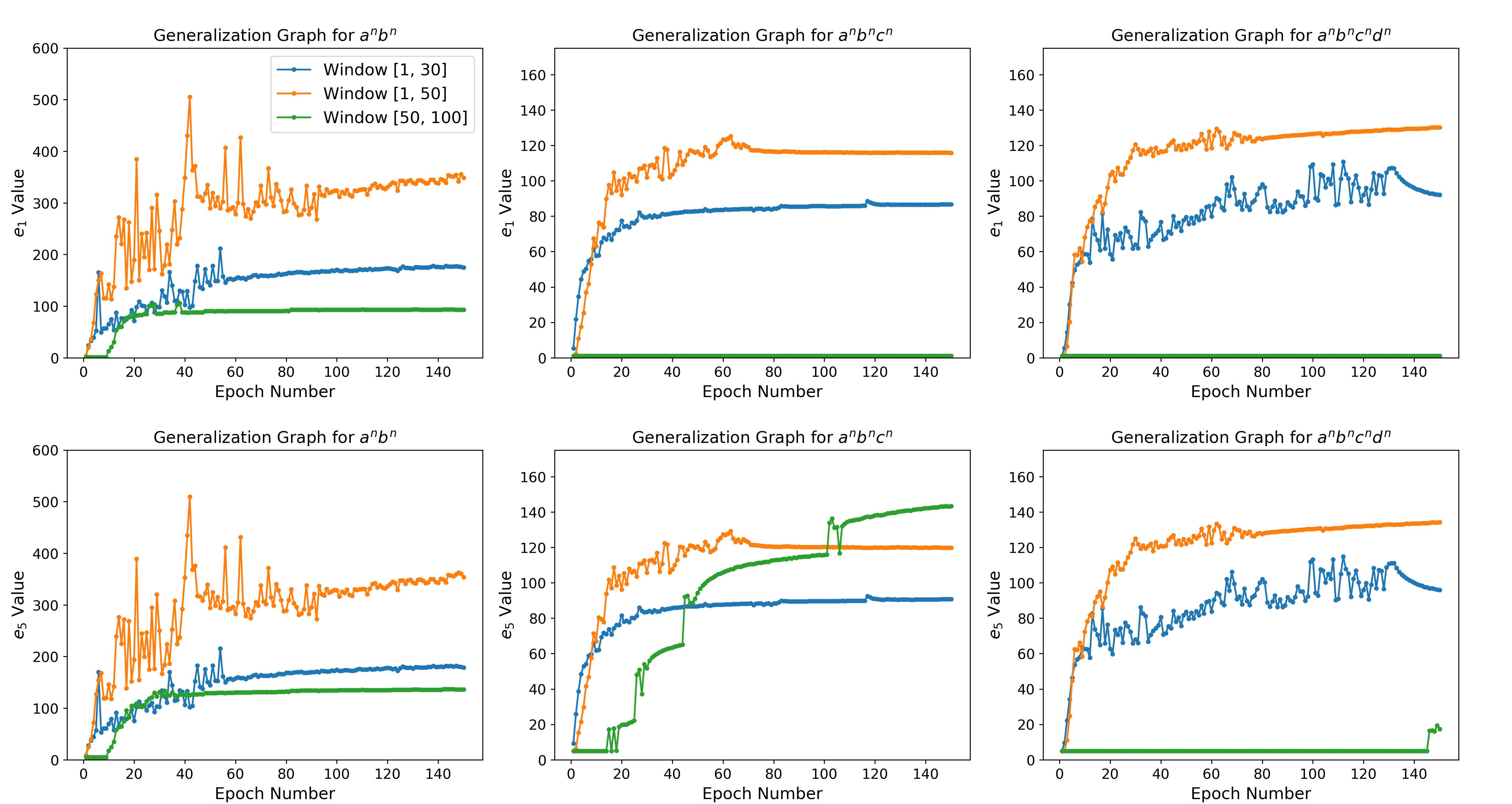}}
%%%%%%%
\caption{Generalization graphs showing the average performance of LSTMs trained under different training length-windows for each language. The top plots show the $e_1$ values, whereas the bottom ones the $e_5$ values.}
\label{fig:windowgraphs}
%\vspace{-5pt}
\end{figure*}

 \subsection{Length Distributions}
Figure \ref{fig:distributiongraphs} exhibits the generalization graphs for the three formal languages trained with LSTM models under different length distribution regimes. Each single-color sequence in a generalization graph shows the average performance of ten LSTMs trained under the same settings but with different weight initializations. In all these experiments, the training sets had the same length-window $[1, 50]$. On the other hand, we used $2$, $3$, and $4$ hidden units in our LSTM architectures for the languages $a^n b^n$, $a^n b^n c^n$, and $a^n b^n c^n d^n$, respectively.\footnote{The results with other configurations were qualitatively similar.} The top three plots show the average lengths of the shortest sequences ($e_1$) whose outputs were incorrectly predicted by the model at test time, whereas the bottom plots show the fifth such shortest lengths ($e_5$). We note that the models trained on uniformly distributed samples seem to perform the best amongst all the four distributions in all the three languages. Furthermore, for the languages $a^n b^n c^n$ and $a^n b^n c^n d^n$, the U-shaped Beta-Binomial distribution appears to help the LSTM models generalize better than the left- and right-tailed Beta Binomial distributions, in which the lengths of the samples are intentionally skewed towards one end of the training length-window. 

When we look at the plots for the $e_1$ values, we observe that all the distribution regimes seem to facilitate learning at least up to the longest sequences in their respective training datasets, drawn by the light blue horizontal lines on the plots, except for the left-tailed Beta-Binomial distribution for which we see errors at lengths shorter than the training length threshold in the languages $a^n b^n c^n$ and $a^n b^n c^n d^n$. For instance, if we were to consider only the $e_1$ values in our analysis, it would be tempting to argue that the model trained under the left-tailed Beta-Binomial distribution regime did not learn to recognize the language $a^n b^n c^n d^n$. By looking at the $e_5$ values, in addition to the $e_1$ values, we however realize that the model was actually learning many of the sequences in the language, but it was just struggling to recognize and correctly predict the outputs of some of the short sequences in the language. This phenomenon can be explained by the under-representation of short sequences in left-tailed Beta-Binomial distributions. Our observation clearly emphasizes the significance of looking beyond $e_1$, the shortest error length at test time, in order to obtain a more complete picture of the model's generalizing capabilities.

 \subsection{Training Length Windows}

Figure \ref{fig:windowgraphs} shows the generalization graphs for the three formal languages trained with %single-layer 
LSTM models under different training windows. 
We note that enlarging the training length-window, naturally, enables an LSTM model to generalize far beyond its training length threshold. Besides, we see that the models with the training length-window of $[50, 100]$ performed slightly better than the other two window ranges in the case of $a^n b^n c^n$ (green line, bottom middle plot). Moreover, we acknowledge the capability of LSTMs to recognize longer sequences, as well as shorter sequences. For instance, when trained on the training length-window $[50, 100]$, our models learned to recognize not only the longer sequences but also the shorter sequences not presented in the training sets for the languages $a^n b^n$ and $a^n b^n c^n$.

Finally, we highlight the importance of the $e_5$ values once again: If we were to consider only the $e_1$ values, for instance, we would not have captured the inductive learning capabilities of the models trained with a length-window of $[50, 100]$ in the case of $a^n b^n c^n$, since the models always failed at recognizing the shortest sequence $ab$ in the language. Yet, considering $e_5$ values helped us evaluate the performance of the LSTM models more accurately.

\begin{figure*}[t]
\centering
{\includegraphics[width=0.99\textwidth]{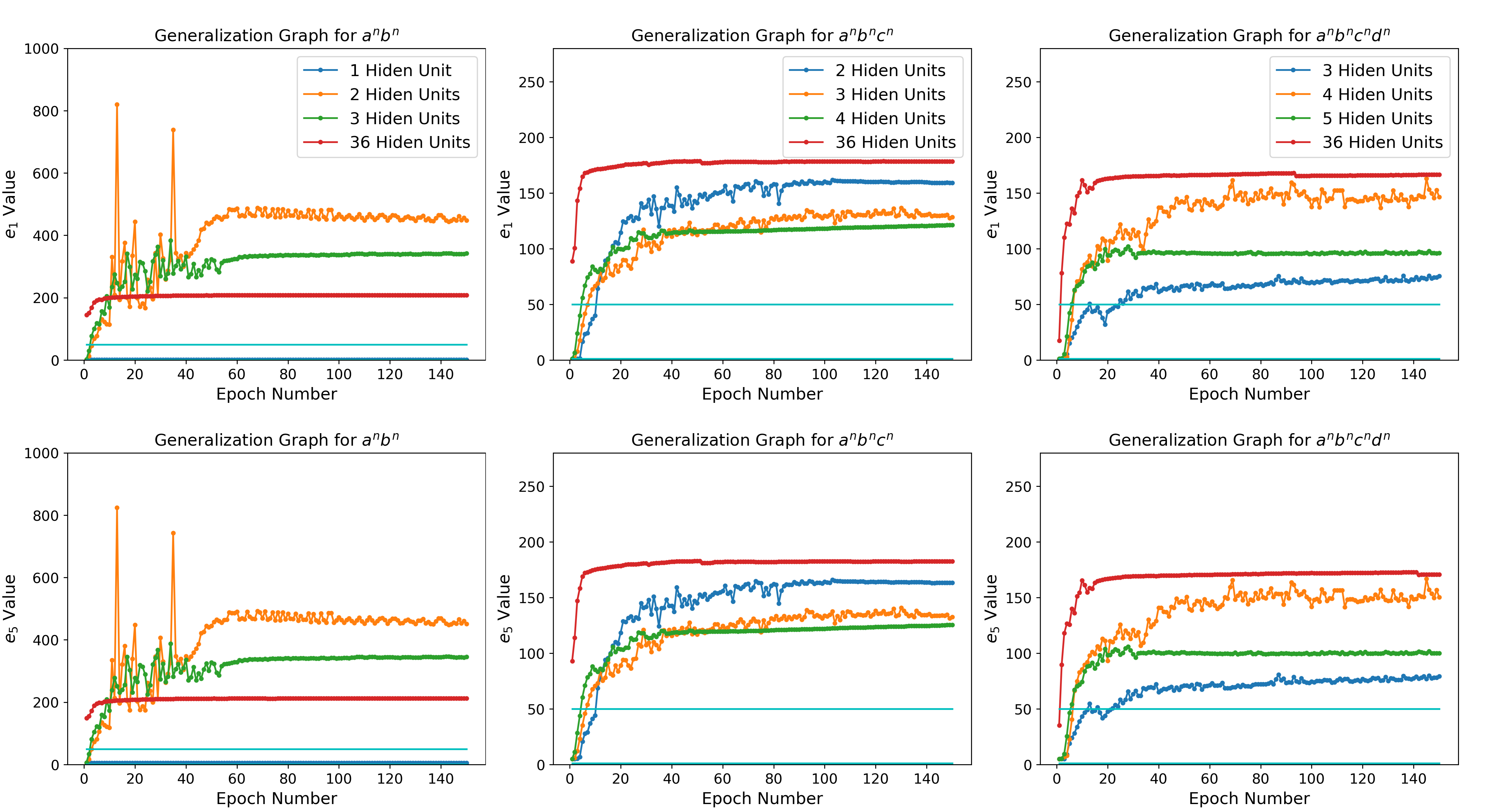}}
%%%%%%%
\caption{Generalization graphs showing the average performance of LSTM models with a different number of hidden units for each language. The top plots show the $e_1$ values, whereas the bottom ones the $e_5$ values. The light blue horizontal lines indicate the training length window $[1, 50]$.}
\label{fig:hiddenunitsgraphs}
\end{figure*}

\subsection{Number of Hidden Units}
There seems to be a positive correlation between the number of hidden units in an LSTM network and its stability while learning a formal language. As Figure \ref{fig:hiddenunitsgraphs} demonstrates, increasing the number of hidden units in an LSTM network both increases the network's stability and also leads to faster convergence. However, it does not necessarily result in a better generalization.\footnote{The results shown in the plot are  for models that were trained on datasets with uniform length distributions with a length window of $[1,50]$. We observed similar trends with other configurations.} We conjecture that, with more hidden units, we simply offer more resources to our LSTM models to regulate their hidden states to learn these languages. The next section supports this hypothesis by visualizing the hidden state activations during sequence processing.

\section{Discussion}
In addition to the analysis of our empirical results in the previous section, we would like to touch upon two important characteristics of LSTM models when they learn formal languages, namely the convergence issue and counting behavior of LSTM models. 

\paragraph{Convergence:}

\begin{figure*}[h]
\centering
{\includegraphics[width=0.49\textwidth]{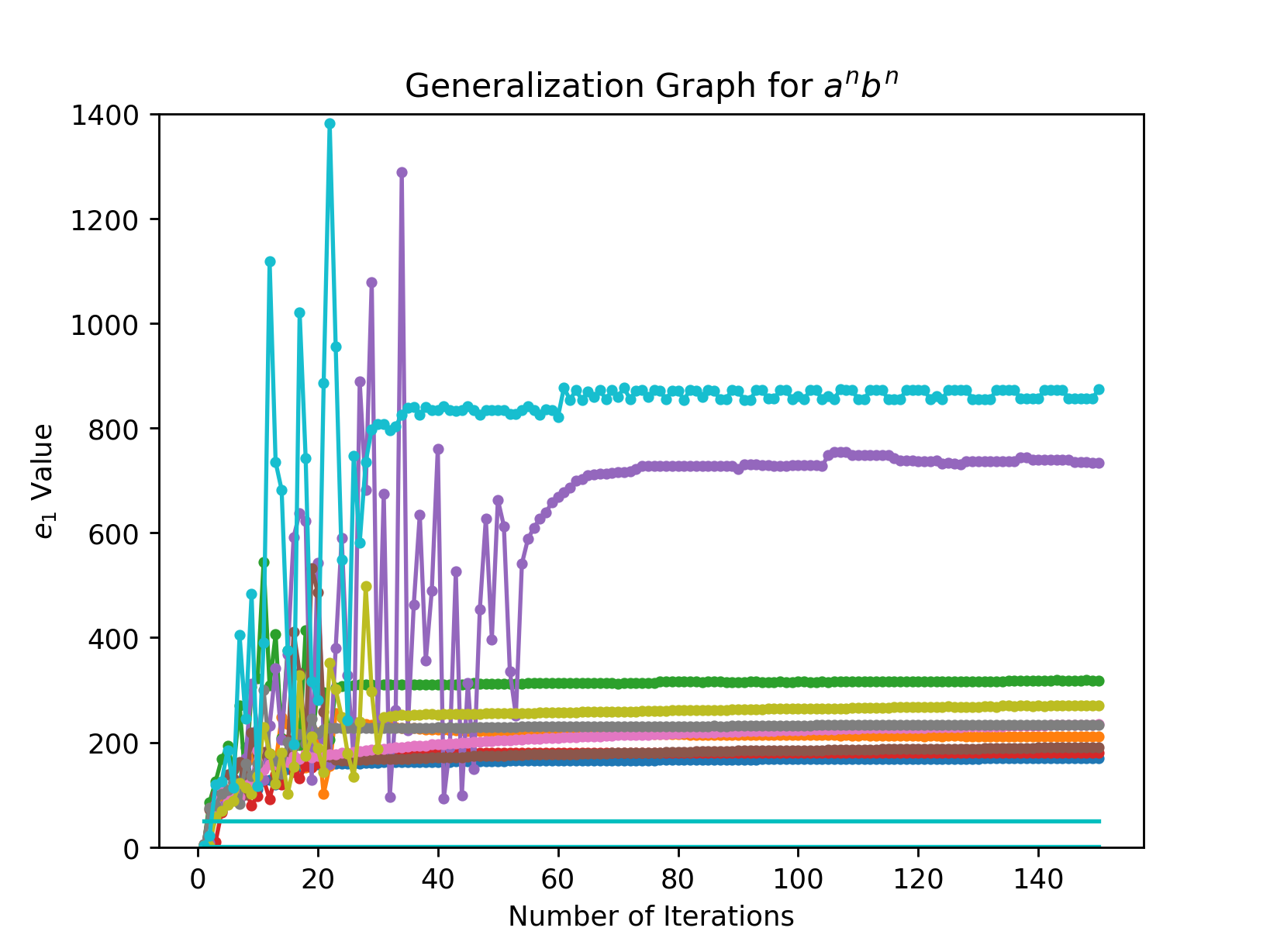}}
{\includegraphics[width=0.49\textwidth]{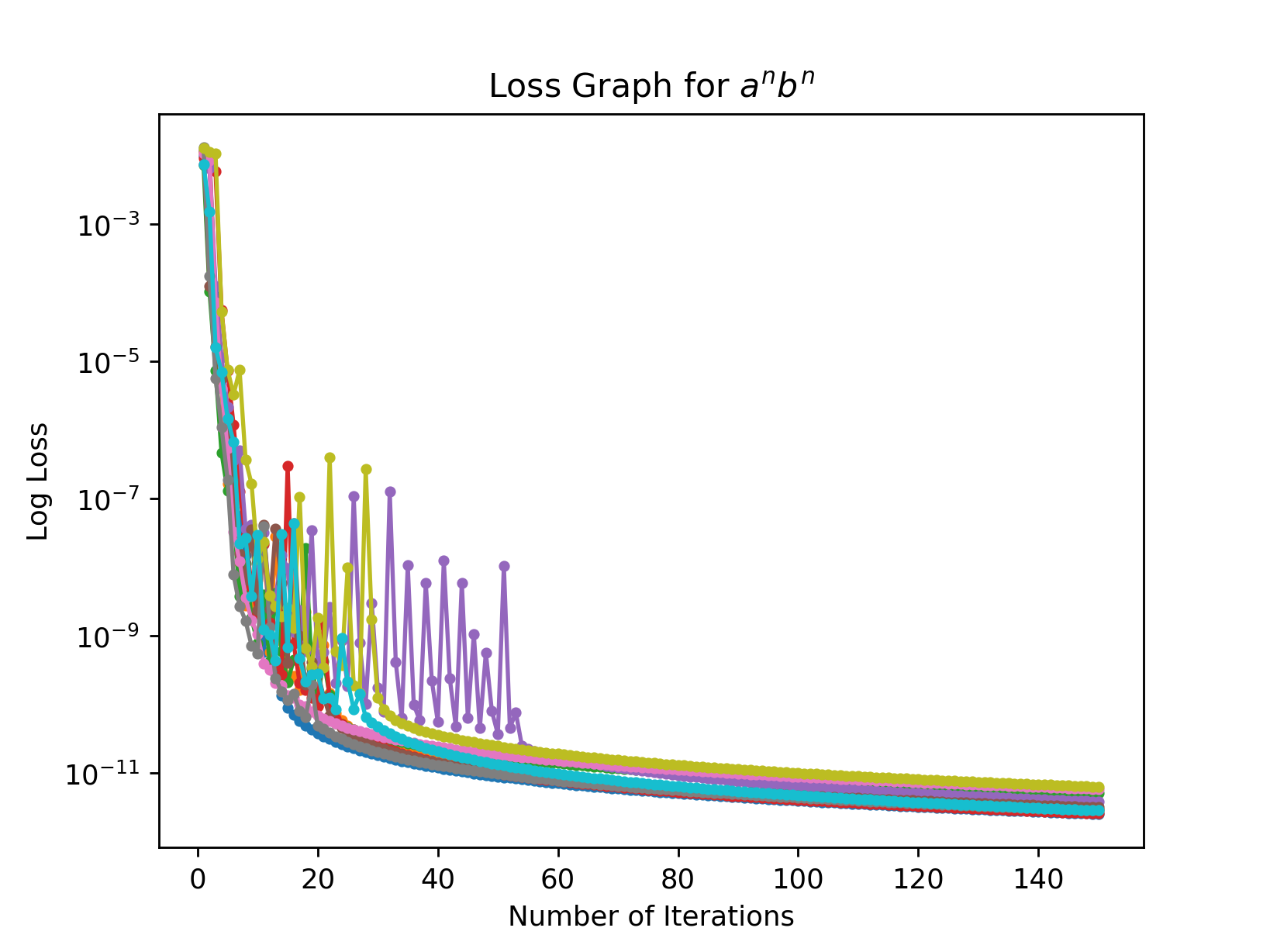}}
\caption{Generalization graph (left) and loss graph (right) with different random weight initializations.}
\label{fig:lossaccuracy}
\vspace{-5pt}
\end{figure*}

We note that our experiments indicate that LSTM models often do not generalize to the same value in a given experiment setting. Figure \ref{fig:lossaccuracy}, for instance, displays the generalization and loss graphs of  LSTM models which were trained to recognize the language $a^n b^n c^n$ under a uniform distribution regime with a training window of $[1, 50]$. The figure shows the results of $10$ trials with different random weight initializations. 
While all runs appear to converge to a similar loss value, they have different generalization values (that is, their $e_1$ values are all different). 
This pattern is fairly common in our experiments, suggesting a disconnection between loss convergence and generalization capability. This result again highlights the importance of performing a fine-grained evaluation of generalization capability, rather than reporting a single number. Our argument is also consistent with those of \citet{boden2002learning} and \citet{chalup2003incremental}, for they also found that the weight initialization affects the inductive capabilities of an LSTM. 

\begin{figure}[t]
\centering
{\includegraphics[width=0.49\textwidth]{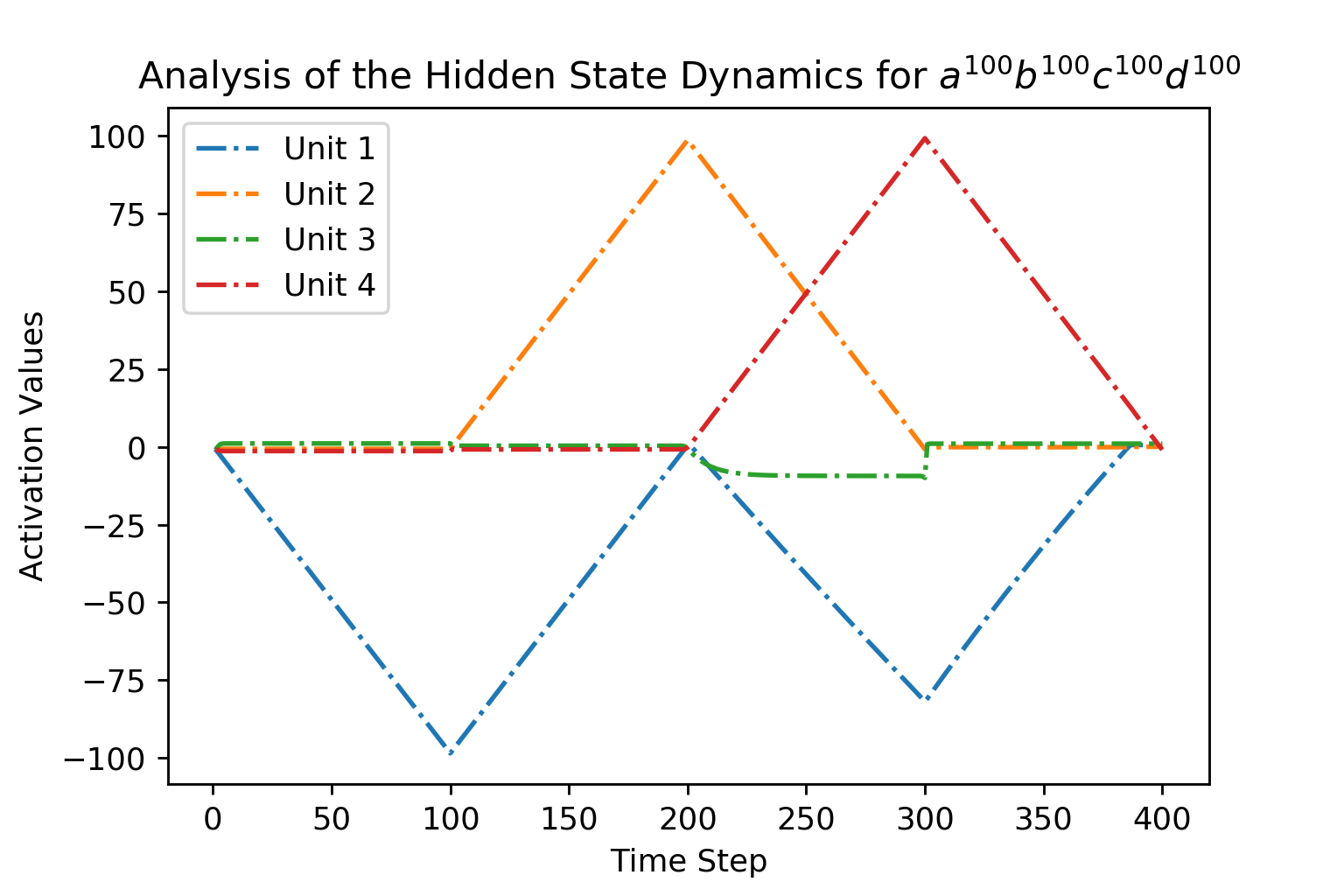}}
\caption{Hidden state dynamics in a four-unit LSTM model. We note that certain units in the LSTM model get activated at time steps $100$, $200$, and $300$.}
\label{fig:learningbycounting}
\end{figure}

\paragraph{Counting Behavior:}

Here we look at the activation dynamics of the  hidden states of the model when processing specific sequences. Figure \ref{fig:learningbycounting} demonstrates that an LSTM network organizes its hidden state structure in such a way that certain hidden state units learn how to count up and down upon the subsequent encounter of some characters. In the case of $a^{100} b^{100} c^{100} d^{100}$, we observe, for instance, that certain units get activated at time steps $100$, $200$, and $300$. In fact, some units appear to cooperate together to count. 

On the other hand, when we visualized the activation dynamics of a model which was trained to learn the language $a^n b^n$ using $36$ hidden units, we observed on the testing of $a^{1000} b^{1000}$ that the %LSTM 
model still uses some of its hidden units to count up and down for all the $a$'s and $b$'s seen by the model, respectively, although it rejects this sample. It simply outputs $(a/b)^{1000}b^{996}\dashv^{4}$, instead of $(a/b)^{1000}b^{999} \dashv$. Our results corroborate and refine the findings of \citet{gers2001lstm} and \citet{P18-2117},
who %also 
noted the existence of a counting mechanisms for simpler languages, while we also observe a collaborative counting behavior in over-parameterized networks.

\section{Conclusion}
In this paper, we have addressed the influence of various length distribution regimes and length-window sizes on the generalizing ability of LSTMs to learn simple context-free and context-sensitive languages, namely $a^n b^n$, $a^n b^n c^n$, and $a^n b^n c^n d^n$. Furthermore, we have discussed the effect of the number of hidden units in LSTM models on the stability of a representation learned by the network: We show that increasing the number of hidden units in an LSTM model improves the stability of the network, but not necessarily the inductive power. Finally, we have exhibited the importance of weight initialization to the convergence of the network: Our results indicate that different hidden weight initializations can yield different convergence values, given that all the other parameters are unchanged. Throughout our analysis, we emphasized the importance of a fine-grained evaluation, considering generalization beyond the first error and during training. We therefore concluded that there are an abundant number of parameters that can influence the inductive ability of an LSTM to learn a formal language and that the notion of \textit{learning}, from a neural network's perspective, should be treated carefully.

\section{Acknowledgment}
The first author gratefully acknowledges the support of the Harvard College Research Program (HCRP) and the Harvard Center for Research on Computation and Society Research Fellowship for Undergraduate Students.
The second author was supported by the Harvard Mind, Brain, and Behavior Initiative. 
The authors also thank Sebastian Gehrmann for his helpful comments and discussion at the beginning of the project. The computations in this paper were run on the Odyssey cluster supported by the FAS Division of Science, Research Computing Group at Harvard University.

\newpage

\bibliography{paper}
\bibliographystyle{acl_natbib}

%\newpage

%\appendix 

\end{document}